\documentclass{article}
\usepackage{spconf,amsmath,graphicx}
\usepackage{subcaption}
\usepackage{amsfonts,epsfig,array,multirow,ltablex,tabularx,setspace,amssymb,multirow}
\usepackage{schemabloc,tikz}
\usepackage{cite}
\usepackage[justification=centering]{caption}
\usepackage[font={small}]{caption}
\usepackage{textcomp}
\usepackage{pifont}
\usepackage{bm}

\title{Statistical Learning of Rational Wavelet Transform for Natural Images}
\name{Naushad Ansari$^\ast$ \thanks{$^\ast$Thanks to CSIR (Council of Scientific \& Industrial Research), Govt. of India for funding.} and Anubha Gupta}
\address{SBILab, Deptt. of ECE, IIIT-Delhi, India}
\begin{document}
%
\maketitle
\begin{abstract}
Motivated with the concept of transform learning and the utility of rational wavelet transform in audio and speech processing, this paper proposes \underbar{R}ational \underbar{W}avelet Transform \underbar{L}earning in \underbar{S}tatistical sense (RWLS) for natural images. The proposed RWLS design is carried out via lifting framework and is shown to have a closed form solution. The efficacy of the learned transform is demonstrated in the application of compressed sensing (CS) based reconstruction. The learned RWLS is observed to perform better than the existing standard dyadic wavelet transforms.
\end{abstract}
\vspace{-1em}
\section{Introduction}
\label{Section for Into}
Transform learning (TL) is an active research area where the sparsifying transform along with the transform domain signal are learned using some constraints for a class of signals. Currently, TL is being used in several applications including image/video denoising and MRI reconstruction \cite{ravishankar2015online,wen2015video,ravishankar2015efficient}. While TL is being used actively, non-convexity of the problem having no closed form solution makes it difficult to solve. Hence, greedy algorithms are used to solve TL problem. 

Among existing transforms, discrete wavelet transform (DWT) is widely used in applications because of its ability of efficient signal representation \cite{mallat1999wavelet}. Non-uniqueness of wavelet basis motivates one to learn wavelet transform in applications. Wavelet transform learning can be viewed as a specific case of transform learning. Since the integer translates of the associated wavelet filters form the basis in $l_2$-space, wavelet transform learning corresponds to learning of wavelet filters.

Generally, dyadic wavelet transform is used that decomposes input signal spectrum into two uniform frequency bands via two-channel filterbank. On the other hand, rational wavelet transform (RWT) provides non-uniform frequency band representation of signal spectrum that is seen to be useful in some applications \cite{blu1996iterated,blu1993iterated}. RWT has also been used in pattern recognition \cite{chertov2010non} and feature extraction \cite{le2014optimized}. Although methods for RWT designs have been presented in the literature \cite{bayram2009frequency, bayram2009overcomplete, baussard2004rational, nguyen2013bi}, designed wavelets are independent of the signal of interest. Recently, a method has been proposed in  \cite{ansari2015lifting} to learn rational wavelet deterministically from a given signal. Since \cite{ansari2015lifting} requires full signal, it cannot be used in inverse problems such as CS-based reconstruction where one does not have access to the full original signal. 

This paper proposes rational wavelet learning for natural images. It has been shown that natural images can be modeled as fractional Brownian motion (fBm) processes in\cite{pentland1984fractal}. fBm processes are Gaussian non-stationary random processes with stationary increments that form a class of statistically self-similar processes \cite{mandelbrot1968fractional} and have been used widely in image processing \cite{stewart1993fractional,chen1989fractal}.

The above discussion of transform learning, flexibility of rational wavelet transform, and modeling of natural images via fBm processes motivates us to learn rational wavelet transform for natural images in statistical sense. Specifically, statistics of a set of natural images are used to propose method for learning separable rational wavelet transform for this class of images. Lifting framework for rational wavelet introduced in \cite{ansari2015lifting} is utilized in the proposed work and is called the RWLS method. The proposed formulation leads to convex problem that can be solved by least squares making the RWLS method computationally efficient. 

Following are the salient contributions of this work:
\vspace{-0.6em}
\begin{enumerate}
\item Statistical learning of rational separable wavelet transform for natural images is proposed.
\vspace{-0.6em}
\item Lifting framework, that is Digital Signal Processing (DSP) hardware friendly, is used in the proposed method making the learned transform easily implementable on hardware.
\vspace{-0.6em}
\item The proposed formulation leads to convex problem unlike conventional TL and can be solved easily.
\vspace{-0.6em}
\item The proposed RWLS is applied in compressed sensing based reconstruction and is observed to perform better than the existing dyadic wavelet transforms.   
\vspace{-0.6em}
\end{enumerate}
\vspace{-1em}
\section{Brief Background}
\label{sec:pagestyle}
\subsection{Lifting in Dyadic Wavelet}
\label{Section for Lifting Theory}
Lifting methodology supports customized wavelet design \cite{daubechies1998factoring}, \cite{sweldens1996lifting}. This design is modular, guarantees perfect reconstruction, and allows non-linear filters to be part of the wavelet structures. A general lifting scheme consists of three steps: Split, Predict, and Update (Refer to Fig. 1). In the \textit{split} step, given input signal is divided into even $x_e[n]$ and odd $x_o[n]$ indexed samples. The corresponding filterbank structure is called as the Lazy wavelet system \cite{sweldens1996lifting} and is converted to the conventional wavelet system using successive predict and update stage filters as shown in Fig. 2 with analysis filters labeled as $G_l(z)=Z\{g_l[n]\}$, $G_h(z)=Z\{g_h[n]\}$ and the synthesis filters as $F_l(z)=Z\{f_l[n]\}$, $F_h(z)=Z\{f_h[n]\}$.    

In the \textit{Predict Lifting} step, odd samples are predicted from the neighboring even samples using the predictor $P \equiv T(z)$ or vice-versa. This step modifies the analysis highpass filter and the synthesis lowpass filter as: 

\begin{equation}
G_{h}^{new}(z) = G_h(z)-G_l(z)T(z^{2}).
 \label{Equation1}
\end{equation}
\begin{equation}
F_{l}^{new}(z) = F_l(z)+F_h(z)T(z^{2}).
 \label{Equation2}
\end{equation}
\begin{figure}[!ht]
\vspace{-1em}
\begin{center}
\includegraphics[scale=0.6, trim =6mm 8mm 6mm 3mm]{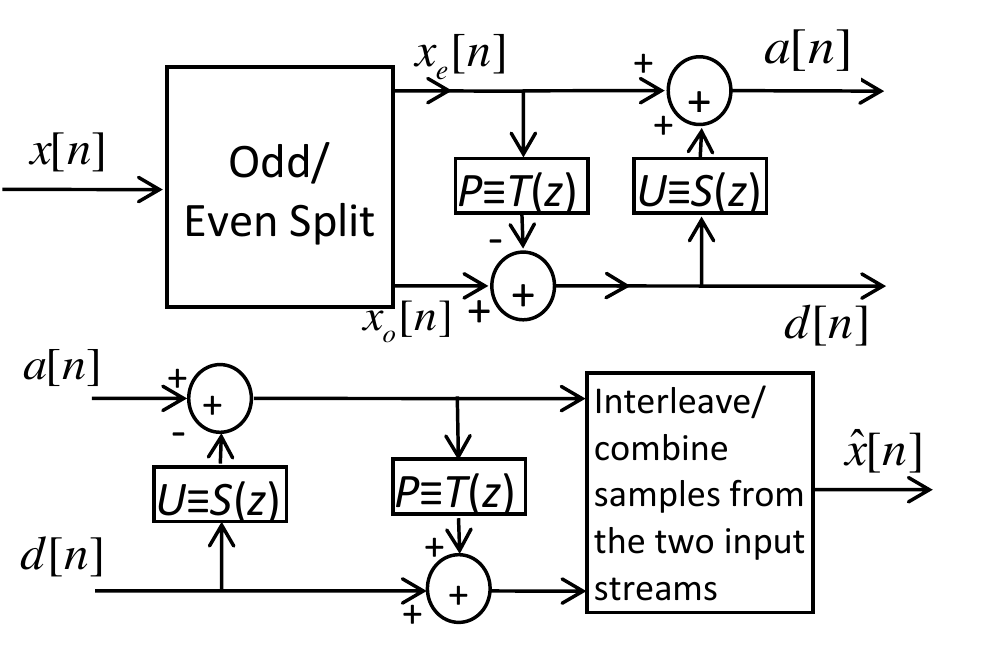}
\end{center}
\caption{Steps of Lifting: Split, Predict and Update}
\vspace{-1em}
\end{figure}
\begin{figure}[!ht]
\begin{center}
\includegraphics[scale=0.6, trim =6mm 6mm 6mm 4mm]{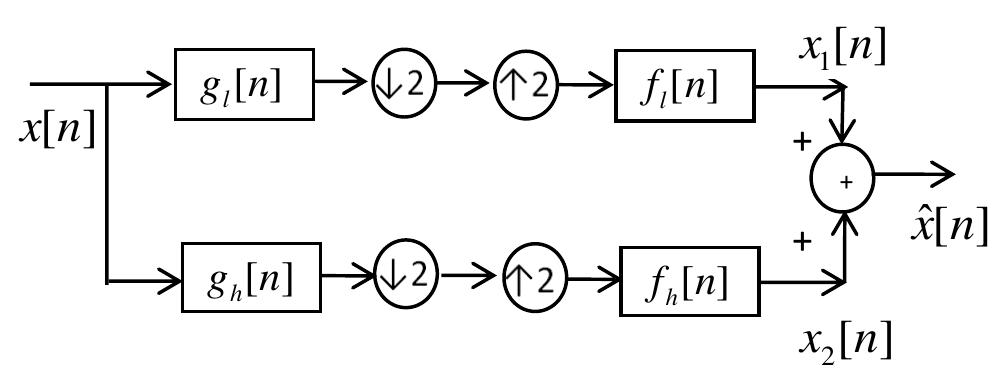}
\caption{Two Channel Wavelet System}
\end{center}
\vspace{-2.0em}
\end{figure} 
The \textit{Update Lifting} step modifies the analysis lowpass filter and the synthesis highpass filter. The update step filter is denoted with the symbol $U \equiv S(z)$ and the related equations are given as: 
\vspace{-0.5em}
\begin{equation}
G_{l}^{new}(z) = G_l(z)+G_h(z)S(z^{2}).
 \label{Equation3}
\end{equation}
\begin{equation}
F_{h}^{new}(z) = F_h(z)-F_l(z)S(z^{2}).
 \label{Equation4}
\end{equation}
\subsection{Rational Wavelet}
\vspace{-0.5em}
Let us consider Fig. 3(a) with 2-channel $(\frac{2}{3},\frac{1}{3})$ rational wavelet filterbank that can be converted into an equivalent uniformly decimated \textit{M}-band structure (Fig. 3(b)). Filters $G_0(z)$ and $G_1(z)$ of Fig. 3(b) can be written as an equivalent filter $G_l(z)$ of Fig. 3(a) using the following equation:
\begin{equation}
G_l(z)=G_0(z^2)+z^3G_1(z^2).
\label{Equation6}
\end{equation}
Similarly, synthesis filters $F_0(z)$ and $F_1(z)$ of Fig. 3(b) can be written as an equivalent filter $F_l(z)$ of Fig. 3(a) using the following equation:
\begin{equation}
F_l(z)=F_0(z^2)+z^{-3}F_1(z^2),
\label{Equation7}
\end{equation}
while other filters remain same, i.e., $G_h(z)=G_2(z)$ and $F_h(z)=F_2(z)$.
\begin{figure}[!ht]
\centering
\begin{subfigure}[b]{0.5\textwidth}
\centering
\captionsetup{justification=centering}
\includegraphics[scale=0.4, trim =4mm 0mm 0mm 4mm]{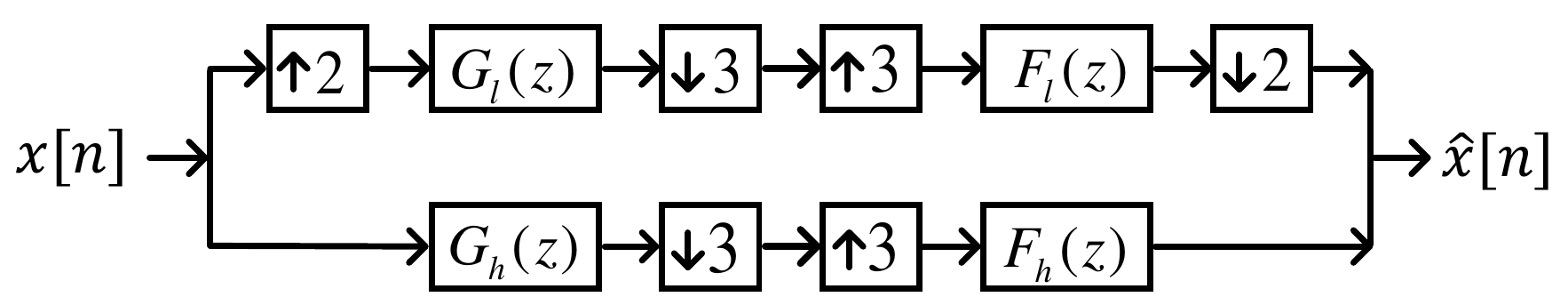}
\caption{Two-channel $(\frac{2}{3},\frac{1}{3})$ rational filterbank}
\vspace{0em}
\end{subfigure} 
\begin{subfigure}[b]{0.5\textwidth}
\centering
\captionsetup{justification=centering}
\includegraphics[scale=0.4]{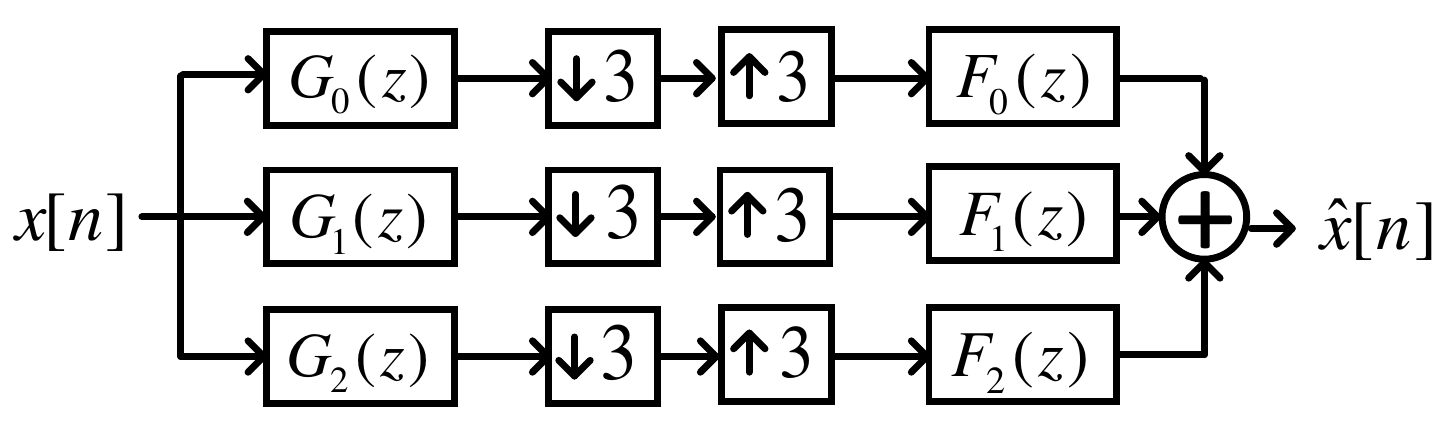}
\vspace{-0.5em}
\caption{Equivalent $M$-band ($M=3$) filterbank}
\end{subfigure}
\vspace{-2em}
\captionsetup{justification=centering}
\caption{Rational wavelet system and its equivalent $M$-band structure}
\vspace{-1.5em}
\end{figure}
\vspace{-1em}
\subsection{Fractional Brownian Motion}
Fractional Brownian motion $B_H(t)$ is a Gaussian, zero mean, self similar, non-stationary random process with stationary increments \cite{manolakis2005statistical}. The auto-covariance of the corresponding discrete time process $B_H[n]$ is given by:
\begin{equation}
r_B^H[n_1,n_2]=\frac{\sigma_H^2}{2}(|n_1|^{2H}-|n_1-n_2|^{2H}+|n_2|^{2H}),
\label{Equation for correlation}
\end{equation}   
where $\sigma_H^2=var(B_H[1])=\frac{1}{\Gamma (2H+1)|sin(\pi H)|}$, and $H$ is the self-similarity index, also called as Hurst exponent. The statistical properties of fBm processes are completely characterized by the single parameter $H$ that can be estimated using the maximum likelihood estimation method presented in \cite{lundahl1986fractional}. 
\vspace{-1em}
\section{Proposed RWLS Learning Method}
This section presents the proposed RWLS learning method on $(\frac{2}{3},\frac{1}{3})$ rational wavelet statistically matched to natural images. Learning of separable two-dimensional (2D) rational wavelet is presented that requires learning 1-D RWLS separately matched to the row space and the column space of natural images. The proposed strategy is identical on either the row or the column space. For the sake of readers' ease, let us first consider design for the column space.
\vspace{-1em}
\subsection{Proposed Learning for the Column Space}
Consider the initial architecture of uniformly decimated 3-band \textit{Lazy} wavelet with filters (in Fig. 3(b)):
\begin{align}
G_i(z)=z^i,\,\,\,\,\,i=0,1,2 \nonumber \\
F_i(z)=z^{-i},\,\,\,\,\,i=0,1,2.
\end{align}
This \textit{Lazy} wavelet is subsequently transformed to equivalent $(\frac{2}{3},\frac{1}{3})$ rational wavelet via \eqref{Equation6} and \eqref{Equation7}. On feeding the vectorized column form of collection of natural images, labeled as $x[n]$, through this rational \textit{Lazy} wavelet filterbank, following approximate $a[n]$ and detail $d[n]$ subband coefficients are obtained:
\begin{align}
 a[n]=
	\begin{cases}
		x\left[\frac{3n}{2}\right] & \quad \text{if } n \text{ is a even}\\
		x\left[\frac{3n-1}{2}\right] & \quad \text{if } n \text{ is a odd},
	\end{cases}
\end{align}
\begin{equation}
d[n]=x[3n+2]
\end{equation}
Next, the lowpass and highpass filters of the \textit{Lazy} rational wavelet structure are lifted via predict and update stage polynomial learned as explained in the following subsections.
\vspace{-1em}
\subsubsection{Predict stage}
\begin{figure}[!ht]
\centering
\includegraphics[scale=0.4, trim=0mm 4mm 0mm 0mm]{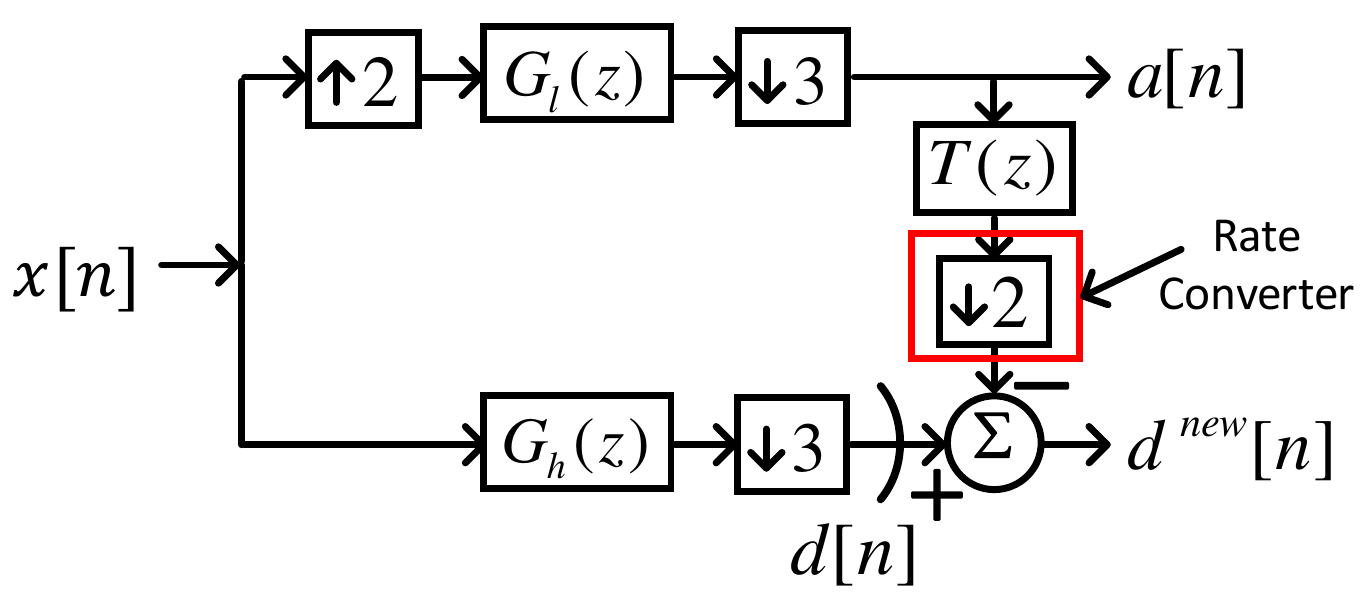}
\vspace{-0.7em}
\caption{Predict stage}
\label{Figure for predict stage}
\vspace{-1em}
\end{figure}

We require to predict one branch of samples with the help of the other branch in the  predict stage. In rational wavelet structure, this requires the concept of rate converter as proposed in \cite{ansari2015lifting} because the output sample rate of two branches is unequal (refer to Fig. \ref{Figure for predict stage}). Considering the predict polynomial filter $T(z)$ as 
\vspace{-0.4em}
\begin{equation}\label{eq:T}
T(z)=t_0z+t_1z^2,
\vspace{-0.8em}
\end{equation} 
we obtain 
\begin{align}\label{dnew}
d^{new}[n]&=d[n]-t_0x[3n+1]-t_1x[3n+3], \nonumber \\
&=x[3n+2]-t_0x[3n+1]-t_1x[3n+3].
\end{align} Thus, the choice of $T(z)$ in (\ref{eq:T}) allows $d[n]=x[3n+2]$ to be exactly predicted from the neighboring samples. These updated detail coefficients $d^{new}[n]$ can also be viewed as the error in predicting the lower branch samples. Hence, (\ref{dnew}) is re-written as
\begin{equation}
e[n]=d^{new}[n]=x[3n+2]-t_0x[3n+1]-t_1x[3n+3].
\end{equation}
$T(z)$ is learned by minimizing the mean squared prediction error (mse) given by:
\begin{equation}\label{mse}
\zeta[n]=E(e^2[n])=E(\{x[3n+2]-t_0x[3n+1]-t_1x[3n+3]\}^2),
\end{equation} where $E(.)$ denotes the expectation operator. 
\begin{figure}[!ht]
\vspace{-0em}
\centering
\includegraphics[scale=0.35]{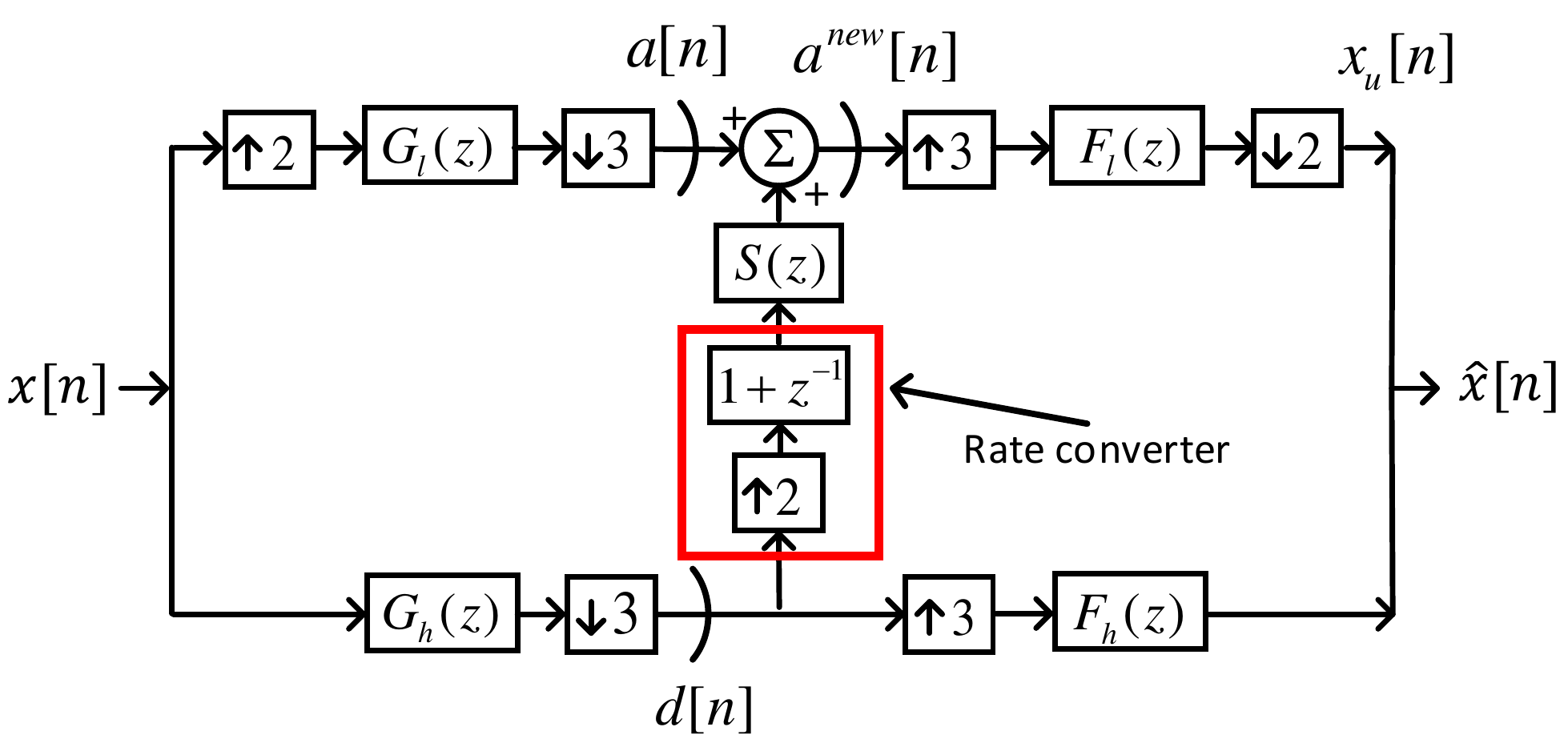}
\vspace{-0.5em}
\caption{Update stage}
\label{Figure for update stage}
\vspace{-1.5em}
\end{figure}
To minimize mse, mse vector $\bm{\zeta}$ is we differentiated with respect to $\mathbf{t}$ and is equated to zero as:
\begin{align}\label{mmse}
\frac{\partial \bm{\zeta}}{\partial \mathbf{t}}=&0-2E[\mathbf{A}'\mathbf{b}]+2E[\mathbf{A}'\mathbf{A}]\mathbf{t}=0 \nonumber \\
\implies  &E[\mathbf{A}'\mathbf{A}]\mathbf{t}=E[\mathbf{A}'\mathbf{b}].
\end{align}
Assuming that the input signal $x[n]$, corresponding to the column space of natural images, belongs to an fBm process, $E[\mathbf{A}'\mathbf{A}]$ and $E[\mathbf{A}'\mathbf{b}]$ are computed using (\ref{Equation for correlation}) and (\ref{mmse}) is solved for $\mathbf{t}$. On simplifying the structure of Fig. 4, the updated equivalent analysis highpass filter, using the learned predict filter $T(z)$, can be written as: 
\vspace{-1em}
\begin{equation}\label{Gh_update}
\vspace{-1em}
G_h^{new}(z)=G_h(z)-\sum_{k=0}^{1}G_l(z^{\frac{1}{2}}W_2^{2k})T(z^{\frac{3}{2}W_2^{3k}}),
\end{equation}
where $W_k=e^{-j\frac{2\pi}{k}}$. 
For the update of the corresponding synthesis lowpass filter, the rational wavelet structure is converted to the equivalent $3$-band structure and the polyphase matrix $\mathbf{E}(z)$ of the analysis side is computed using $G_0(z)$, $G_1(z)$ and $G_2^{new}(z)$. On applying the condition of perfect reconstruction \cite{vaidyanathan1993multirate} in \eqref{Equation for PR}, polyphase matrix $\mathbf{R}(z)$ of the synthesis side is computed.
\vspace{-0.5em}
\begin{equation}
\vspace{-0.5em}
\textbf{R}(z)\textbf{E}(z)=cz^{-n_0}\textbf{I},
\label{Equation for PR}
\end{equation}
where $c \in \mathbb{R}$, $n_0 \in \mathbb{Z}$, and $\mathbf{I}$ is $3 \times 3$ identity matrix. From $\mathbf{R}(z)$ and \eqref{Equation7}, updated filter $F_1^{new}(z)$ of the rational wavelet is computed. This completes the predict stage.
\subsubsection{Update Stage}
\begin{figure*}[t]
\centering
\begin{subfigure}[b]{0.08\textwidth}
\includegraphics[scale=0.08]{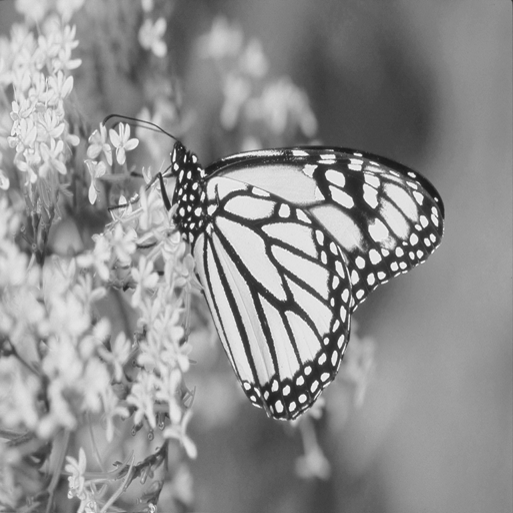}
\caption*{Img1}
\label{Img1}
\end{subfigure}
\begin{subfigure}[b]{0.08\textwidth}
\includegraphics[scale=0.08]{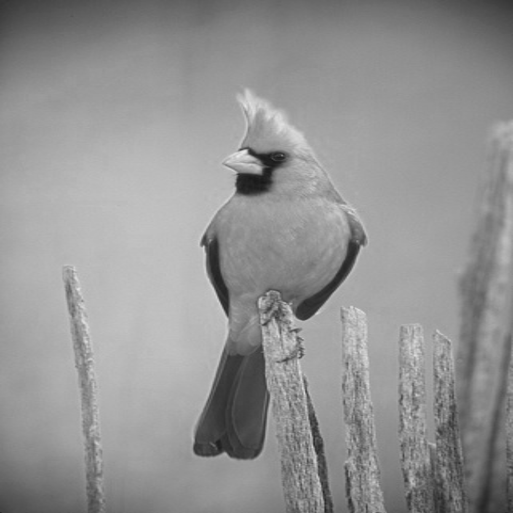}
\caption*{Img2}
\label{Img2}
\end{subfigure}
\begin{subfigure}[b]{0.08\textwidth}
\includegraphics[scale=0.08]{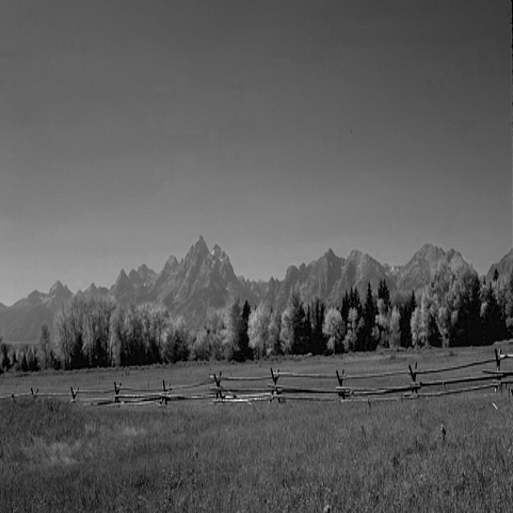}
\caption*{Img3}
\label{Img3}
\end{subfigure}
\begin{subfigure}[b]{0.08\textwidth}
\includegraphics[scale=0.08]{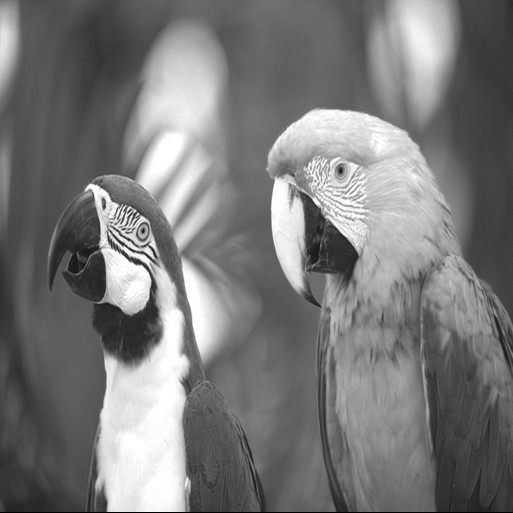}
\caption*{Img4}
\label{Img4}
\end{subfigure}
\begin{subfigure}[b]{0.08\textwidth}
\includegraphics[scale=0.08]{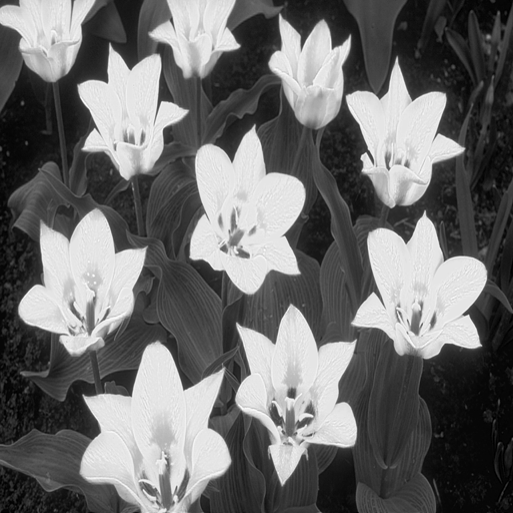}
\caption*{Img5}
\label{Img5}
\end{subfigure}
\begin{subfigure}[b]{0.08\textwidth}
\includegraphics[scale=0.08]{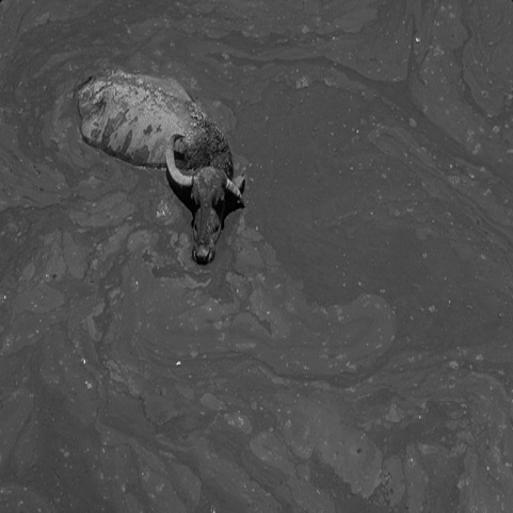}
\caption*{Img6}
\label{Img6}
\end{subfigure}
\begin{subfigure}[b]{0.08\textwidth}
\includegraphics[scale=0.08]{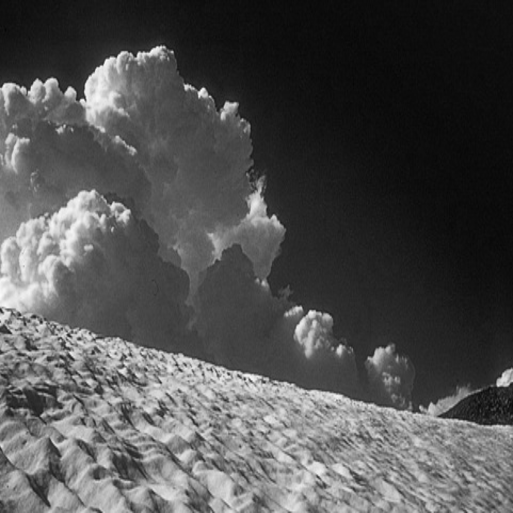}
\caption*{Img7}
\label{Img7}
\end{subfigure}
\begin{subfigure}[b]{0.08\textwidth}
\includegraphics[scale=0.08]{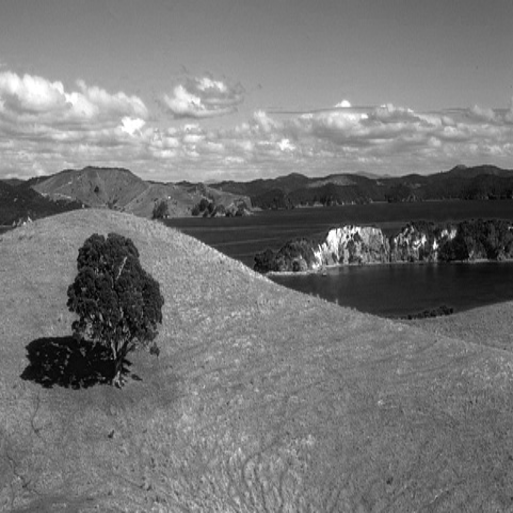}
\caption*{Img8}
\label{Img8}
\end{subfigure}
\begin{subfigure}[b]{0.08\textwidth}
\includegraphics[scale=0.08]{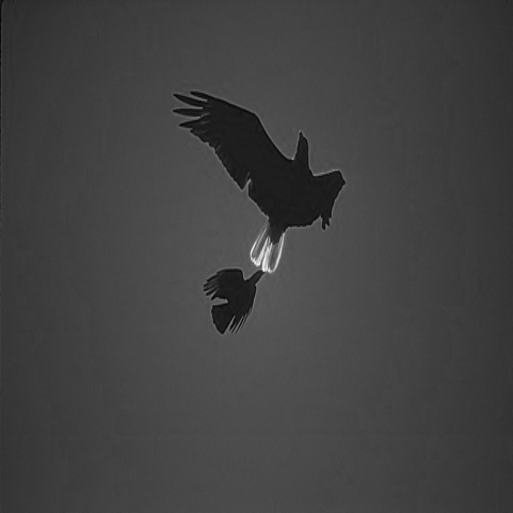}
\caption*{Img9}
\label{Img9}
\end{subfigure}
\begin{subfigure}[b]{0.08\textwidth}
\includegraphics[scale=0.08]{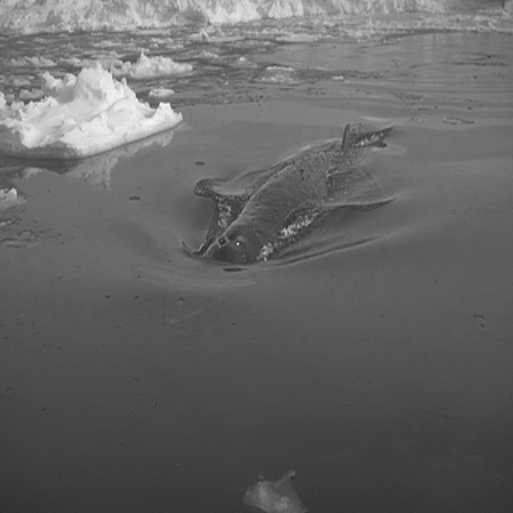}
\caption*{Img10}
\label{Img10}
\end{subfigure}
\begin{subfigure}[b]{0.08\textwidth}
\includegraphics[scale=0.08]{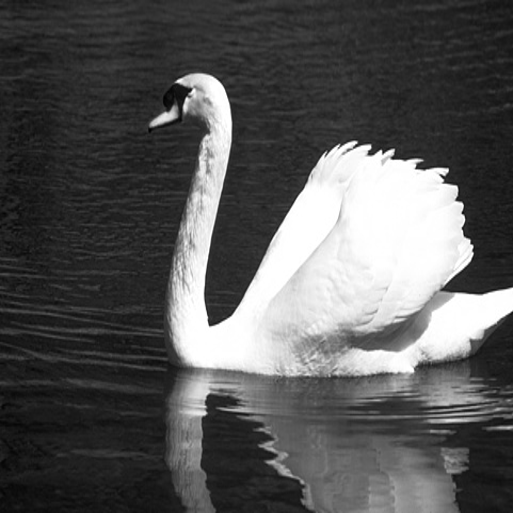}
\caption*{Img11}
\label{Img11}
\end{subfigure}
\vspace{-0.8em}
\captionsetup{justification=centering}
\caption{Images used in the experiment. Img1 to Img10 were used to design the matched rational wavelet system and the performance in CS reconstruction is tested on images Img1, Img4 and Img11.}
\label{Fig for All Images}
\vspace{-1.0em}
\end{figure*}
Next, the update stage filter $S(z)$ shown in Fig.5 is learned. Again, rate converter, shown in Fig.5 as proposed in \cite{ansari2015lifting}, is required. The reconstructed signal at the upper branch is shown as $x_u[n]$. Since the natural images are generally rich in low frequency content, $x_u[n]$ should be as close as possible to the input signal $x[n]$. This allows us to learn the update stage filter by minimizing the energy difference of the two signals as below:
\vspace{-1em}
\begin{equation}\label{update_eq}
\tilde{\mathbf{s}}=\min_{\mathbf{s}} ||\mathbf{x}-\mathbf{x}_u||^2,
\vspace{-1em}
\end{equation}   
where $\textbf{s} \equiv S(z)=s_0+s_1z^{-2}$. Signal $\mathbf{x}_u$ can be written in terms of update stage filter \textbf{s} that allows us to solve (\ref{update_eq}). Once $S(z)$ is learned, analysis lowpass filter is updated as:
\begin{equation}
G_l^{new}(z)=G_l(z)+G_h(z^2)S(z^3)
\end{equation}
Synthesis highpass filter is updated using the method similar to the one used to update the synthesis lowpass filter.
This completes the proposed learning. Since the lifting framework is modular, more predict and update stages can be appended to get longer length filterbanks. This is to note that for learning the RWT for the column space of natural images, we vectorized an ensemble of natural images column-wise and stacked them below each other to build a 1-D signal. Next, we estimate the Hurst exponent \textit{H} of this column vector and learn the RWT as presented above.
\vspace{-0.8em} 
\subsection{Proposed Design for the Row Space}
\vspace{-0.8em}
Corresponding to the row-space design, we vectorize all images row-wise and stack them to build a 1-D signal. Next, we estimate the Hurst exponent \textit{H} of this row vector and learn the  RWT using the method presented in the previous sub-section.
\section{Application}
\begin{figure}
\centering
\begin{subfigure}[b]{0.49\textwidth}
\centering
\includegraphics[scale=0.4]{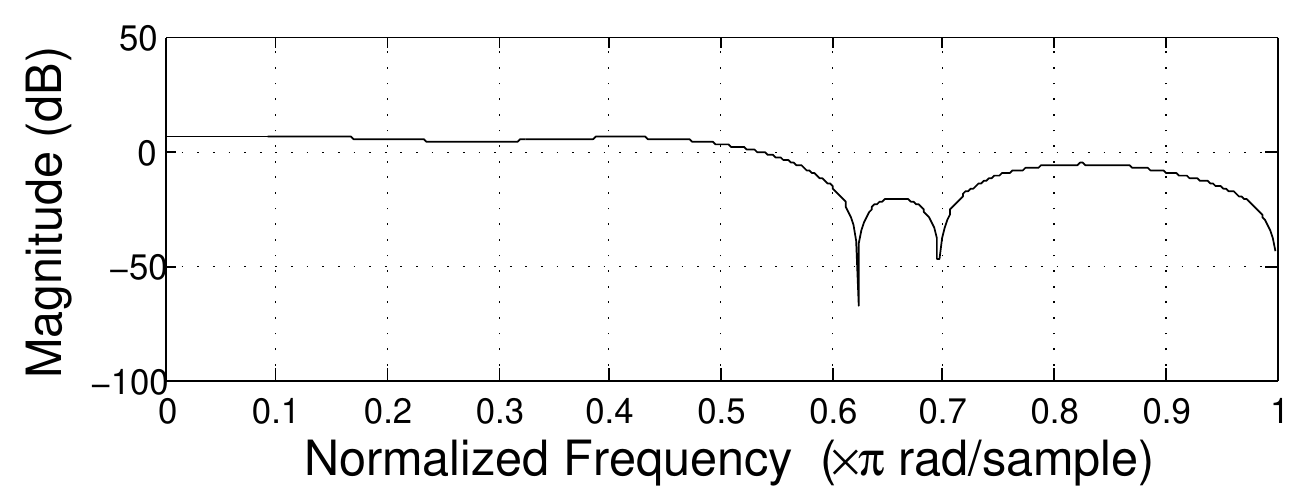}
\vspace{-0.5em}
\caption{Low pass filter}
\end{subfigure}
\begin{subfigure}[b]{0.49\textwidth}
\centering
\includegraphics[scale=0.4]{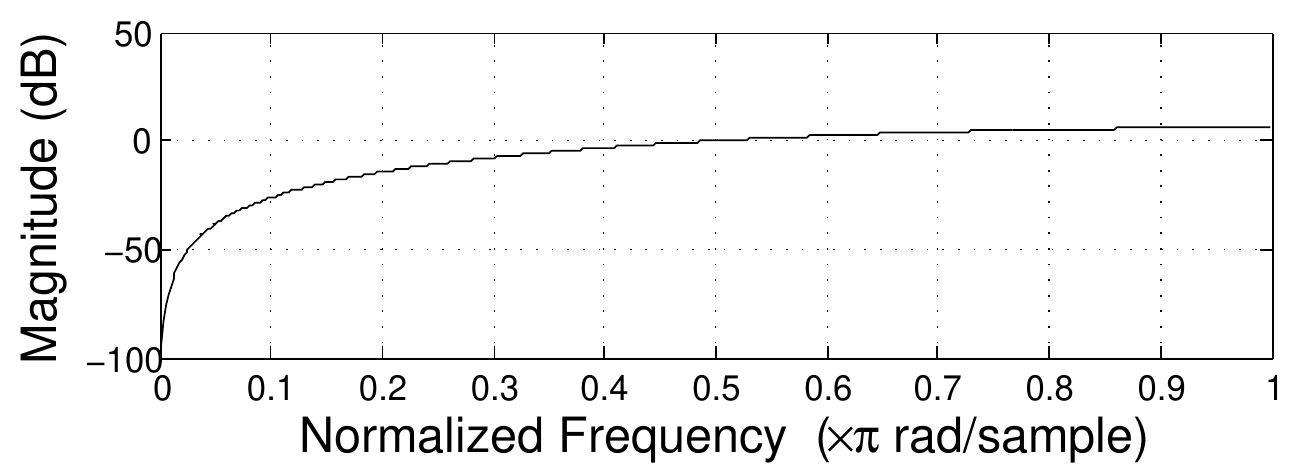}
\vspace{-0.5em}
\caption{High pass filter}
\end{subfigure}
\vspace{-2em}
\caption{Frequency response of analysis side filters matched to the column space of natural images.} 
\vspace{-1em}
\label{Figure for frequency response}
\end{figure}

The proposed RWLS method is applied on natural images as separable wavelets. The performance of the learned RWLS is compared with standard bi-orthogonal 5/3 and 9/7 wavelets in the application of compressive sensing based reconstruction of natural images of dimension $512 \times 512$. An ensemble of ten natural images shown in Fig. 6 is considered for learning the statistically matched rational wavelet structure for the row space and the column space of images. The value of Hurst exponent is observed to be between 0.5 to 1.0 for all the ten images considered. Fig. \ref{Figure for frequency response} shows the frequency response of the analysis side lowpass and highpass filters matched to the column space of natural images. 

Bernoulli measurement matrix with entries taken as $\pm 1$ is considered in CS. Since it is computationally expensive to apply CS on big images, we use the concept of block CS \cite{gan2007block}, where block-size of $32 \times 32$ is considered. Recently, multilevel wavelet decomposition has been proposed over L-shaped pyramid (L-Pyramid) (Fig. \ref{fig:L-pyramid}) in \cite{ansari2017image} and is observed to perform better in CS application compared to the existing multilevel regular pyramid (R-Pyramid) wavelet decomposition (Fig. \ref{fig:R-pyramid}). We decompose our input images to 3-level using this new L-Pyramid wavelet decomposition in our experiments. Table-I presents reconstruction results in terms of PSNR (peak signal to noise ratio) for sampling ratios varying from $90\%$ to $30\%$, where sampling ratio is the percentage of total samples measured.
\begin{figure}[!ht]
\centering
\includegraphics[scale=0.9]{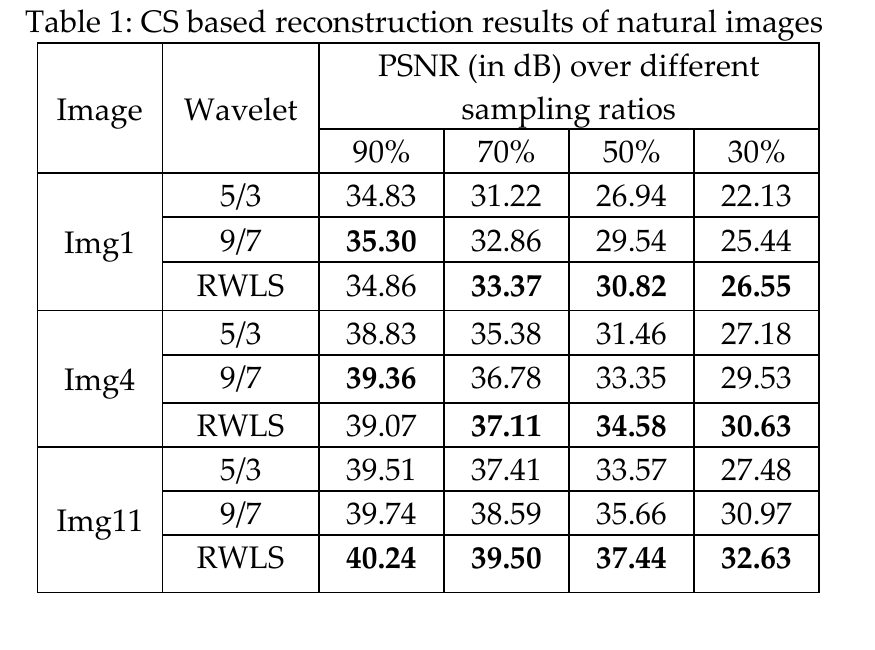}
\vspace{-2.0em}
\end{figure}
\begin{figure}[!ht]
\begin{subfigure}[b]{0.235\textwidth}
\centering
\includegraphics[scale=0.45]{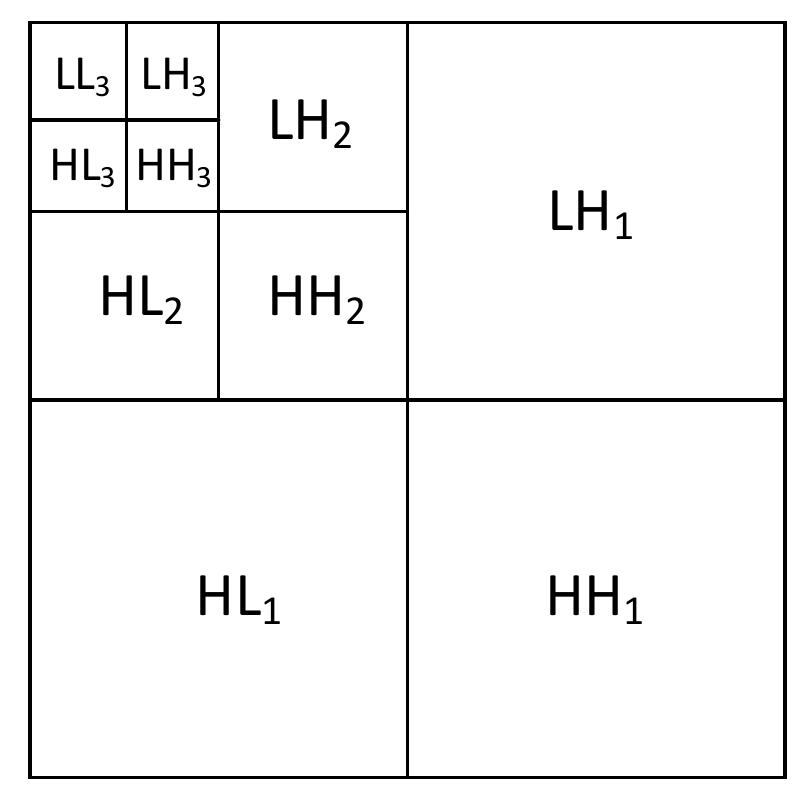}
\caption{\small 3-level R-Pyramid Wavelet Decomposition}
\label{fig:R-pyramid}
\end{subfigure}
\begin{subfigure}[b]{0.235\textwidth}
\centering
\includegraphics[scale=0.45]{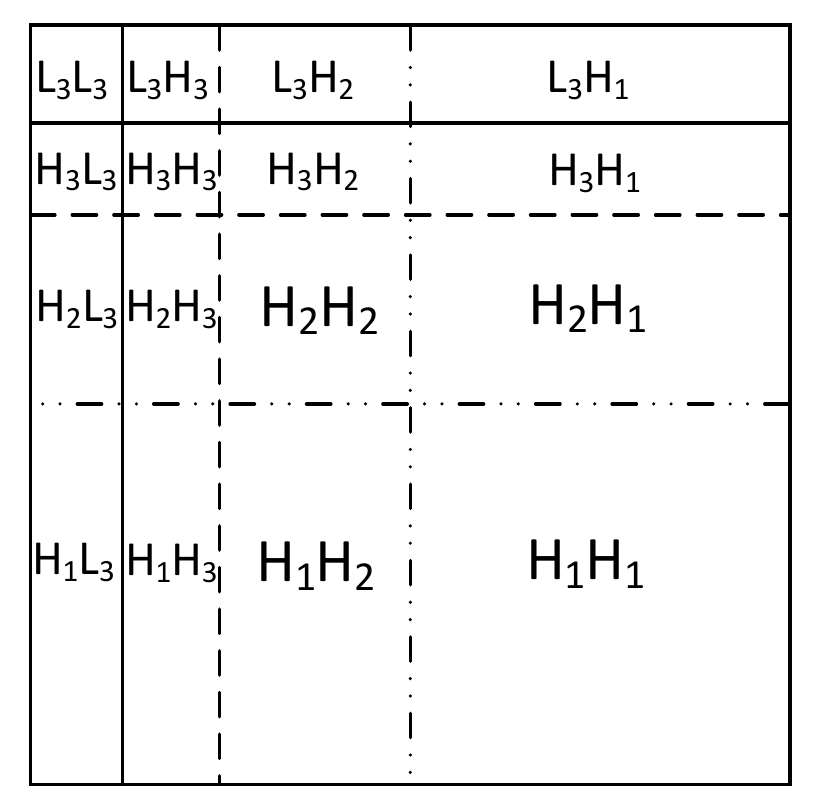}
\caption{\small 3-level L-Pyramid Wavelet Decomposition}
\label{fig:L-pyramid}
\end{subfigure}
\vspace{-1em}
\caption{\small 3-level Wavelet Decomposition \\ Notations for (b): First Letter- wavelet band  of columns (\underbar{L}owpass or \underbar{H}ighpass); Second Letter-wavelet band of rows (\underbar{L}owpass or \underbar{H}ighpass); Subscripts 1, 2, and 3 denote the number of times wavelet has been applied on that row/column} 
\vspace{-1em}
\label{Figure for wavelet pyramid}
\end{figure}

From Table-1, we note that the performance of the proposed RWLS is superior (comparable at 90\% for Img1 and Img4) to standard wavelets on natural images. Although image `Img11' was not used in the ensemble of images used to learn the RWT, the performance of the learned RWLS over this image is also superior indicating that the proposed learning indeed provides statistically-matched rational system for the class of natural images.

\vspace{-0.8em}
\section{Conclusion}
\vspace{-0.8em}
Statistical learning for rational wavelet transform (RWLS) method for natural images is presented in this work. The natural images are modeled as fBm processes and their statistical properties are used to learn separable rational wavelet transform. Lifting framework for the rational wavelet is used in the proposed work that provides closed form solution for learning making the method computationally efficient. The learned rational wavelet transform is tested in the application of CS based reconstruction of natural images and is observed to perform better compared to the existing standard bi-orthogonal wavelet transforms.   
\bibliographystyle{IEEEtran}
\bibliography{references}

\end{document}